# Boosting Mask R-CNN Performance for long, thin Forensic Traces with Pre-Segmentation and IoU Region Merging

Moritz Zink,
AIUI GmbH,
Ilmenau, Germany,
moritz.zink@ai-ui.ai

Martin Schiele,
AIUI GmbH,
Ilmenau, Germany,
martin.schiele@ai-ui.ai

Pengcheng Fan,
AIUI GmbH
Ilmenau, Germany,
pengcheng.fan@ai-ui.ai

Stephan Gasterstädt
AIUI GmbH
Ilmenau, Germany
stephan.gasterstaedt@ai-ui.ai

The High-throughput and Unified Toolkit for Trace Analysis by Forensic Laboratories in Europe - SHUTTLE for short, (https://www.shuttle-pcp.eu/) is a project funded by the EU under Horizon 2020. The goal is to develop a robot that scans and analyzes specific crime scene traces in a fully automated manner. The project has an initial duration of 48 months, consists of three phases and was funded with 10.5 million euros. The goal is to facilitate the work of laboratories through the use of AI and automation technology. The robot developed by the consortium "AG SHUTTLE Toolkit Jena" is able to scan, classify and then segment an A4-sized surface with a resolution of 2.5 µm in less than 5 hours. The amount of data is enormous. With a "field of view" of 17.7 x 13.3 mm and a total area of 210 x 297 mm (A4), around 51 gigabytes of data are generated per image (~150 megapixels, 16 bits á 300 megabytes memory requirement) and image channel. During each run, 91 channels (RGB Reflected Light HiRes 3 channels, RGB Transmitted Light HiRes 3 channels, Fluorescence HiRes 3 channels, Spectrum without Polarization LoRes 41 channels, Spectrum with Polarization LoRes 41 channels), are measured as standard, resulting in more than 300 GB of data to be processed.

*Abstract*—Mask R-CNN has recently achieved great success in the field of instance segmentation. However, weaknesses of the algorithm have been repeatedly pointed out as well, especially in the segmentation of long, sparse objects whose orientation is not exclusively horizontal or vertical. We present here an approach that significantly improves the performance of the algorithm by first pre-segmenting the images with a PSPNet algorithm. To further improve its prediction, we have developed our own cost functions and heuristics in the form of training strategies, which can prevent so-called (early) overfitting and achieve a more targeted convergence. Furthermore, due to the high variance of the images, especially for PSPNet, we aimed to develop strategies for a high robustness and generalization, which are also presented here.

## I. Introduction

Unlike traditional Semantic Segmentation, Instance Segmentation allows individual examples of the same class to be represented in a segmented fashion, even if they overlap. Mask R-CNN was originally developed by FAIR (Facebook AI Research Group) and is the industry standard for this task due to its outstanding results for this purpose [1]. However, it has been repeatedly shown that the use of Mask R-CNN in segmenting long, thin objects with diagonal orientation does not lead to the desired success. The mask formed (Fig. 1) is interrupted or incomplete and does not adequately represent the object to be segmented.

This can be attributed to several causes:

On the one hand, both in application and in applied research, the methodology of transfer learning is used almost exclusively. Here, the models are pre-trained, i.e. without randomized initialized weightings. If the dataset used for pre-training and the dataset to be trained are very divergent, an adapted training strategy is necessary to achieve the desired result, especially when using complex models with many hidden layers [2].

Furthermore, current implementations of state-of-the-art deep learning models have an immense number of hyperparameters that need to be optimized in order to cover the respective use case with sufficient accuracy. Even with modern optimization methods, such as Optuna [3], the dimensionality is so high that it is hardly possible to achieve meaningful results in finite time.

Furthermore, a specified adaptation of the hyperparameters of the model implementation to a single problem, such as the recognition of long, thin objects, can lead to a deterioration of the already satisfactory result of the segmentation of other object classes.

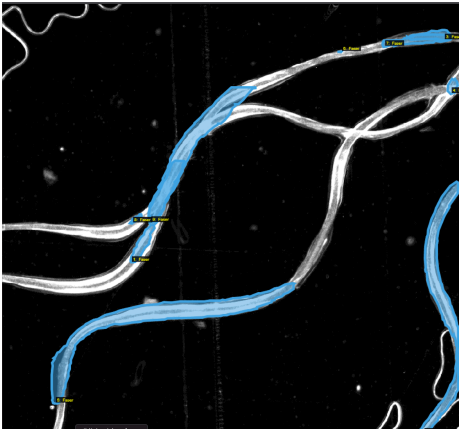

Abb. 1: Incompletely formed masks by Mask R-CNN on a 1024x1024 fiber darkfield microscopy.

## II. Data Inherencies

The images on which this project is based are complex in several respects. They are forensic traces, such as blood and glass, but also fibers. In order to be able to image very small traces, as well as their different inherent attributes and details, a high resolution of 1024x1024 pixels is essential. Furthermore, for the exact representation of all trace attributes the RGB color space must be used. Furthermore, the objects to be recognized are quite different among each other, especially in the class "fibers" (cf. Fig. 2 and Fig. 3). This leads to a further complication in the recognition of the correct class and demands an extremely good generalization of the model (1), i.e. in the first step an empirical risk as low as possible:

$$R(\widehat{f_n}) = \frac{1}{N} \sum L\left(Y, \widehat{f_n}(X)\right), mit \quad (1)$$

$$\lim_{N \to \infty} \to \mathbb{E}\left[L\left(Y, \widehat{f_n}(X)\right) \Big| Tn\right]$$

where: $(X,Y) \sim \mathbb{P}^{(X,Y)}$ independent of von $Tn$, but unknown. For the best possible generalization, [4] must hold:

$$\mathbb{E}\left[L\left(Y, \widehat{f_n}(X)\right) \Big| Tn\right] - \frac{1}{N} \sum L\left(Y, \widehat{f_n}(X)\right) = 0 \quad (2)$$

Due to the unknown probability distribution, equation (2) cannot be calculated. Rather, this theoretical consideration leads to an expression about the (calculable) probabilities:

$$P\left(\mathbb{E}R(\widehat{f_n^*}) - \mathbb{E}R(\widehat{f_n}) \leq \varepsilon\right) \geq 1 - \delta \quad (3)$$

The tolerance limits fort he parameters $\varepsilon$ and $\delta$ are difficult to dertermine theoretically. Qualitatively, however, the generelization can be determined as follows: Condition (3) is sufficiently fulfilled exactly if the prediction always remains constant by selectively removing training examples from the data set, which is what we are trying to achieve here.

In addition, the data already show that the variance within a class varies considerably, which in turn must be represented by the model. In technical terms, this means: A high variance of the pixels $x_i$ on all channels (4) leads to more representable information within the image and facilitates the correct classification based on the more pronounced feature appearance.

$$Var(x_1) = \sigma^2 = \frac{1}{N} \sum (x_i - \overline{x_1})^2 \quad (4)$$

Or more practically, the higher the contrast in the image, the easier it is for the model to correctly represent the desired class, but the more different the contrast is within the same classes, the more difficult it is to make an exact prediction within this class (cf. Fig. 2 and Fig. 3).

In addition, correct classification is further complicated by the image acquisition process: Forensic traces are usually masked on a microscope slide with a special type of transparent adhesive tape [5]. This creates air bubbles between the actual traces on the image (see marking, Fig. 2, for an example). This becomes particularly clear when the trace image is compared with the actual annotations (Fig. 4).

The aforementioned air bubbles have similar structures to elements from the class of "fibers", which further favors misclassification, and consequently incorrect segmentation, on the part of the model used.

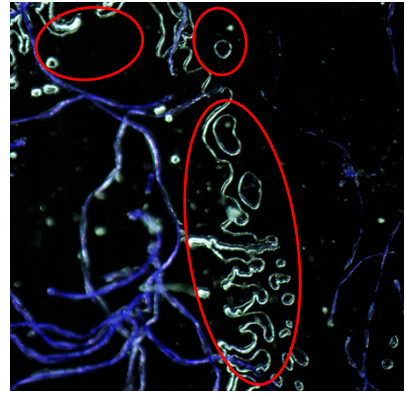

Fig. 2: Randomized selected fiber example 1

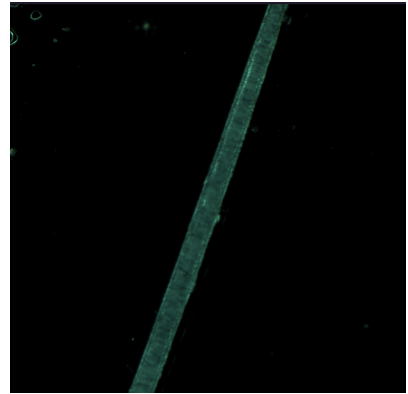

Fig. 3: Randomized selected fiber example 2

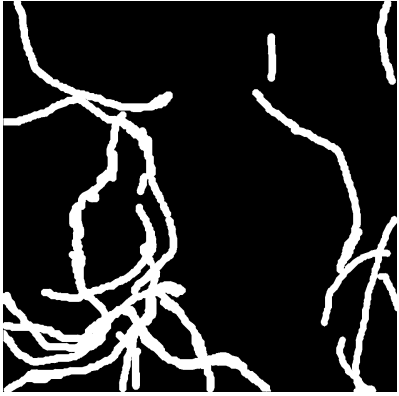

Fig. 4: Annotation image for fiber example 1 (cf. Fig. 1)

## III. Presegmentation with PSPnet

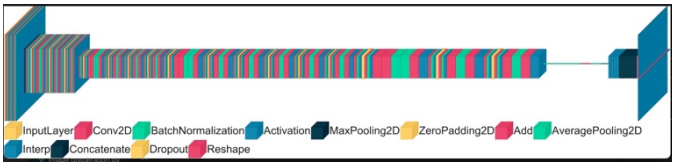

Fig. 5: PSPNet Architecture

TABLE I. MEASURES TO IMPROVE GENERALIZATION

| Amound of loss function of the test data | | |
|---|---|---|
| *Maßnahme* | *Effective* | *Not effective* |
| Increasing dropout to 0.4 | x | |
| ReLu to Swish in deeper layers | x | |
| Use heuristics for learning rate | x | |
| Deeper ResNet Backbone | | x |
| Geometric augmentations | x | |
| Additional non geometric augmentations | x | |
| „Dice Entropy" | o | o |
| Using smaller bachtsizes | x | |

The Pyramid Scene Parsing Network was introduced during the ImageNet Challenge 2016 and is designed for semantic segmentation of complex environments [6].Figure 5 shows the architecture we used (Fig. 5). It is clear that this is an encoder structure, where features are first extracted from the image of dimension $\mathbb{R}^{1024 \times 1024 \times 3}$ and then expanded back into an image of the same basic dimension, i.e. R $\mathbb{R}^{1024 \times 1024}$. However, this is not realized by an "ordinary" decoder, but by a convolutional layer followed by upsampling.

The images are now pre-segmented using the presented model so that only the fiber class remains visible. Any air bubbles, dust inclusions or foreign classes should be removed from the image. This can avoid a false positive representation of foreign objects as fibers, which can occur especially during the transition in the image from the fiber to the air bubble.

Since we require very high robustness for our use case due to the diversity of the class "fibers", several small changes were made within the model structure, which further increase the required high generalization. Table 1 shows the actions taken regarding the model, the training methodology and the data preparation (Tab. 1). Since we are not able to measure the generalization error directly, we use test metrics (see Section D, (7)) for evaluation. These can be implicitly considered as measures of generalization if they were computed on previously unseen data, which is why a division into "train", "validation" and "test data" was made at the beginning. Due to the structure and diversity of the fiber data, a test set size of 20% of the total data set has been chosen, whereas the validation data accounts for 16% and the training data for the remaining part. Thus, it is possible to ensure a representative sample of all fiber subcategories in sufficient quantity.

In the following, we will first take a closer look at the regularization measures tested in Table 1.

### A. Regularization measures within the model

When using dropout, randomly selected neurons and their weights are excluded during the training process and not provided with an adjustment. This results in a simulation of ensemble learning, but with much less (time) effort. The diversity of the networks during training is exponentially correlated to the network depth and width. During a prediction there is no more "dropout", so the prediction is averaged over the different trained models [7]. This results in smaller weightings, which in turn leads to an additional regularization effect.

Even though the use of higher dropout rates between convolutional layers of a model is considered little studied [2] or even controversial, we still used this to reduce/prevent overfitting later in the training process and to stabilize convergence. (Fig. 6). For this purpose, the dropout layer was used after the encoder part of the network.

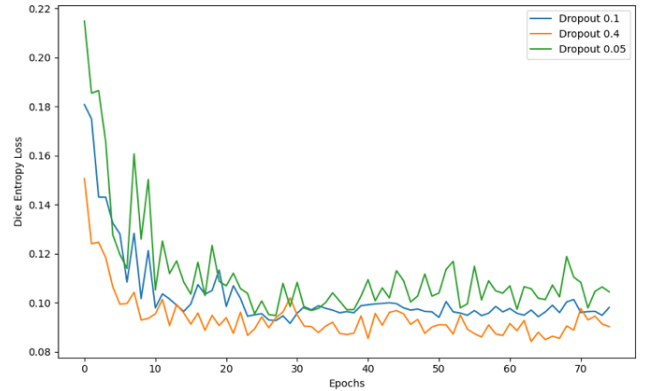

Fig. 6: Different Dropoutrates

### B. Regularization through Data Augmentation

Data augmentation is also one of the regularization techniques that can be used to avoid overfitting. Since there are countless augmentation techniques, we have made the following division:

- No Augmentations
- Geometric Augmentations consisting of resizing, adding rotation and distortions

- Geometric data augmentation additionally with Gaussian Blur, Gaussian Noise, brightness adjustment and sharpness filter, called "advanced augmentations"

It could be shown that by using extended data augmentation techniques, an increased stability in the training process becomes visible. A tendency to overfitting is further reduced and the evaluation metrics turn out better compared to a pure geometric augmentation (Fig. 7 and Fig. 8). Due to the randomly high value of the loss function in the middle of the training, a graph has been inserted which underlines the facts once again.

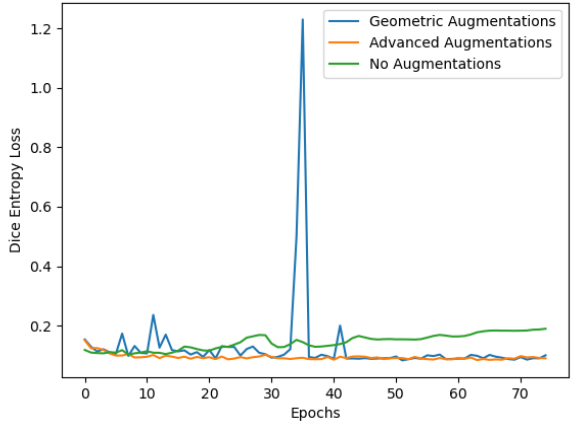

Fig. 7: Overall training progression of different augmentation methodologies.

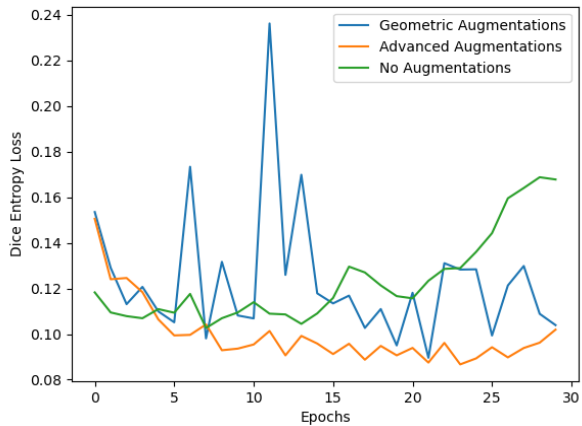

Fig. 8: First half of the training course of different augmentation methodologies.

## C. Use of Heuristics for the Learning Rate

The use of heuristics, especially in the field of deep learning, is gaining more and more attention in research and thus leads to a better theoretical explainability than purely empirically designed experiments [9]. In those it could already be shown that the use of learning rate decay yields particularly good results [10].

General recommendations for an application of the "correct" learning rate are [11]:

1: First, use a high learning rate to avoid unwanted local minima and to speed up the training, especially at the beginning.

2: Subsequently, the learning rate should be reduced to avoid oscillations and allow for more targeted convergence

In principle, the relationship from (5) applies to the error function in neural networks:

$$L: D \subset \mathbb{R}^n \to \mathbb{R} \qquad (5)$$

But for this function $L$ to be convex, the following must hold for the Hessian matrix of them (6):

$$H_L(\boldsymbol{w}) = \left(\frac{\partial^2 L}{\partial w_i \partial w_j}(\boldsymbol{w})\right) \geq 0_{, i,j=1\ldots n} \qquad (6)$$

*However, this is generally not the case in neural networks.*

Thus, the (empirically) derived rules of previous works are obsolete and cannot be confirmed, especially for deeper neural networks. This observation is also made in other works [12]. Here the core statements are related to non-convex, deeper models:

1: An initial high learning rate prevents learning noisy data points from the dataset

2: A consecutive decay in the learning rate helps in model adaptation of complex patterns.

Again, we cannot fully support these observations with sufficiently large variance in the data set (and, of course, sufficient training samples), especially using transfer learning and sufficiently large data augmentation.

An initially too high learning rate led to training instabilities and an "early overfitting". Also in the later course, no convergence could be achieved, neither with a constant learning rate nor with an exponentially decreasing learning rate. Based on these findings, a learning rate function has been developed, which is divided into three areas (Fig. 9):

1: - Warmup

2: - Constant Plateau

3: - Descending Course

At this point, it remains to be said that a drop in the learning rate in our experiments was able to achieve better convergence. The question of whether a decreasing learning rate can also find more complex patterns within the data cannot be answered conclusively in our experiments. In addition, however, it becomes apparent that an initially high learning rate does not necessarily lead to the desired result.

In the context of transfer learning, a layer-specific learning rate is also often discussed, but this was not tested further here due to the high model complexity and the very good results with our approach. Also the often cited approach of "fine tuning" [2] was not considered here due to the large difference between the pre-trained dataset and the fiber images.

This is due to investigations that the selection of the layers to be trained always has to be determined empirically with a high expenditure of time and that this selection is strongly model and data dependent and can therefore only rarely be described as worthwhile. The methodology of training the entire model with a sensitive approach seems to be more efficient here.

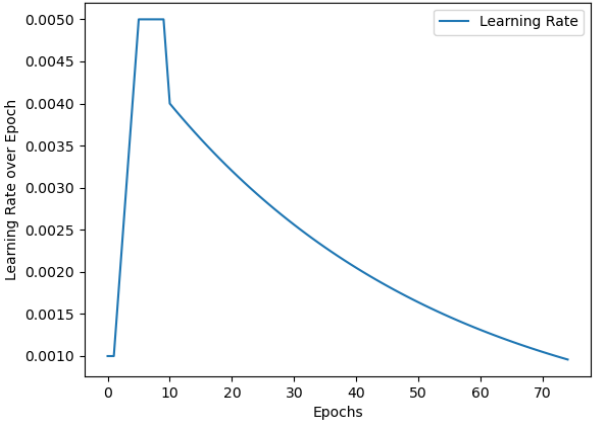

Fig. 9: Learning Rate Function

### D. Verwendung kleinerer Batchsize

The hyperparameter of the batch size is just as important as that of the learning rate. Although the phenomenon is not fully understood, it is often shown empirically that the generalization ability of a model becomes higher the smaller the batch size is chosen, in the extreme case by the value 1. However, this procedure requires an extremely small learning rate to avoid training instabilities due to the noisy gradient. As a result, more training steps are required due to the decreased learning rate and the fact that more time steps must elapse to observe the entire dataset [13]. This is especially evident in previous practical experiments, where the so-called "generalization gap" can be as high as 5%, measured on test metrics. Particularly noteworthy is the fact that this can occur even with an equivalent test cost function. The reason for this can be well explained graphically: optimization methods using large batches tend to generate many positive eigenvalues in $\Delta^2 f(x)$, implying a sharp minimum. However, smaller batches tend to generate a flat minimum in $\Delta^2 f(x)$, due to small eigenvalues [14], (Fig. 10). It is conceptually clear from the figure that a shift between the train and test functions implies a much larger deviation in the cost function for a sharp minimum than for a flat one.

However, since a larger batch size can also have advantages, such as improved computational parallelism and less noisy gradients, we decided to test some batch sizes against each other, starting from the academic example of batch size = 1. The results in terms of Mean IoU on test data are shown in Table 2 (Tab. 2), where the best value from five runs has been presented here.

The Mean IoU is defined as follows (6) and can be understood as a measure for the overlap of the segmentation and the object to be segmented in relation to the sum of both, where A represents the true pixel value and B the segmentation output and applies: $A, B \subseteq \mathbb{S} \in \mathbb{R}^n$ (7) [16]:

$$IoU = \frac{|A \cap B|}{|A \cup B|} \quad (7)$$

The average intersection over union is then derived as follows (8), where N reflects the number of classes and holds true $N \in \mathbb{N}$:

$$Mean\ IoU = \frac{1}{N}\sum_{i=1}^{N} IoU_N \quad (8)$$

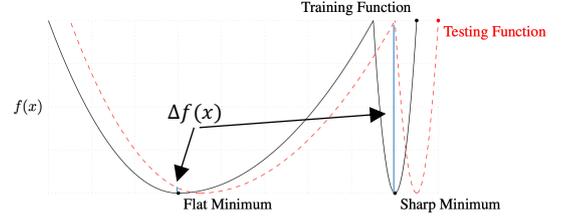

Fig. 10: Cost function with sharp and flat minima, after [14].

TABLE II. CORRELATION MEAN IOU /BATCHSIZE

| Batchsize | MeanIoU |
|---|---|
| Batchsize = 1 | 0.473 ±0.05 |
| Batchsize = 2 | 0.776 ±0.03 |
| Batchsize = 3 | 0.810 ±0.02 |

### E. Costfunction "Dice Entropy"

A typical metric for evaluating the result of a semantic segmentation task is the so-called "Dice Coefficient", where A represents the true pixel value and B the segmentation output on the part of the model (7) (after [15]). There is a clear similarity to the IoU (6), but our goal was to make the test metric as independent from the cost function as possible.:

$$DC = \frac{2|A \cap B|}{|A|+|B|} \quad (9)$$

Since only a ratio of correctly covered pixels is generated here, a cost function can be derived using (8), where ε serves to safeguard borderline cases (such as $A = B = 0$) and $\varepsilon \ll 1$ holds. For a more "practical" implementation, this then yields [15]:

$$L_{DC} = 1 - \frac{2AB + \varepsilon}{A + B + \varepsilon} \quad (10)$$

Equally common in segmentation tasks, as well as in classification problems, is the use of the cost function of "crossentropy", which, depending on the nature of the problem (binary or higher dimensional), is known as "binary-" or "categorical crossentropy". Since we find here only two classes to be segmented, "background" and "fiber", the binary special case of the crossentropy function can be used (9) [15]:

$$L(y, \hat{y}) = -(y \log(\hat{y}) + (1 - y) \log(1 - \hat{y})) \quad (11)$$

Both functions are standard for the task of image segmentation and yet have different operation. In our experiments, it could be shown that the use of a novel cost function may give better results. For this, first (9) has to be extended to the general case for any number of classes (10):

$$L = -\sum_{i=1}^{m} \hat{y}_i \log(p_i) \quad (12)$$

A simple addition of (8) and (10) now leads to our new cost function, which we call "Dice Entropy" (11):

$$L = 1 - \frac{2AB+\varepsilon}{A+B+\varepsilon} + \left(-\sum_{i=1}^{m} \hat{y}_i \log(p_i)\right) \quad (13)$$

The results can now be compared with each other at the same starting weights and are shown in Table 3 (Tab. 3). Again, the Mean IoU was used as metric, also to ensure a certain independence within the validation.

*While segmentation of several classes shows a clear improvement, our binary fiber example rather shows a slight deterioration (Tab. 3).* The values were averaged over five runs, but there is (irrelevantly) greater stability when our cost function is applied to the problem.

TABLE III. EFFECT OF THE COSTFUNCTIONS ON THE SEGMENTATION RESULT

| Costfunction | MeanIoU |
|---|---|
| Dice Entropy | 0.798 ± 0.002 |
| Crossentropy (Binary) | 0.810 ± 0.03 |

*F. Use of a deeper Resnet-Backbone*

Furthermore, it was also investigated how the use of a different Resnet backbone affects the result.

Normally, it can be assumed that the use of a deeper resnet in the encoder structure can provide better results. However, this could not be shown in our experiments. In the depicted training course, the tendency to overfitting is already apparent after a short time. Of course, this is due to the increased number of training parameters, whereby even stronger regularization measures did not lead to success.

The results are shown in Table 4 (Tab. 4) and are also illustrated as a plot of the loss function (Fig. 11).

TABLE IV. INFLUENCE OF THE RESNET BACKBONES

| Resnet | MeanIoU |
|---|---|
| Resnet50 | 0.810 ± 0.03 |
| Resnet101 | 0.721 ± 0.009 |

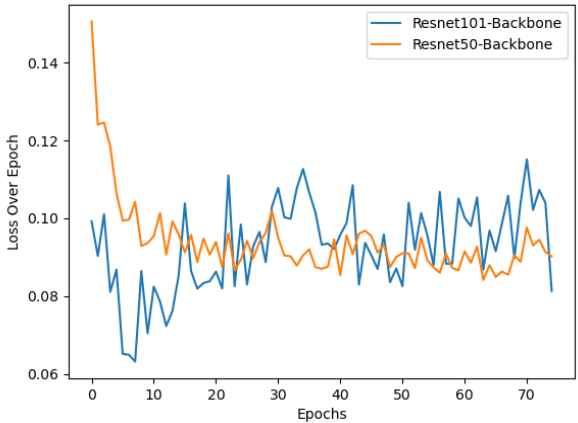

Fig. 11: History of the cost function with different resnet backbones

*G. Swish*

Swish is a quite modern activation feature developed by Google, which is defined as follows [17]:

$$f(x) = x\,\sigma(x) \quad (14)$$

In which:

$$\sigma(x) = \frac{1}{1+e^{-x}} \quad (15)$$

Due to its structure, i.e., non-monotonic and smooth, there are some advantages in its use over ReLu, especially in deeper models. Here ReLu is defined as follows:

$$f(x) = \max(0, x) \quad (16)$$

Activation functions support a convergence of the model mainly by their unrestrictedness upwards. This avoids that the gradient takes very small values, and the training time increases strongly. However, a downward restriction is often desired due to the regularization effect. Both properties can be mapped by the "Swish" function. We have therefore decided to replace the ReLu function by "Swish" in the relevant layers of the model.

The results from our experimental setup are shown below (Tab. 5):

TABLE V. INFLUENCE OF THE ACTIVATIONFUNCTION

| Costfunction | MeanIoU |
|---|---|
| ReLu Baselinemodel | 0.810 ± 0.01 |
| Swish Model | 0.829 ± 0.007 |

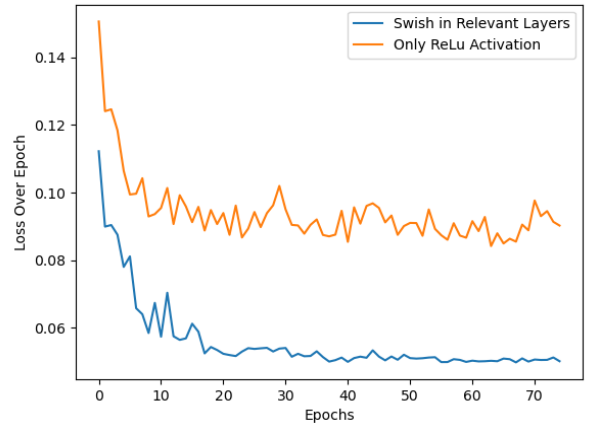

Fig. 12: History of the cost function with Swish instead of ReLu

IV. RESULTS

The following figures show the results of the semantic segmentation of PSPNet. For this purpose, different fiber images were selected in order to emphasize the divergence within the fiber class and the required generalizability of the model.

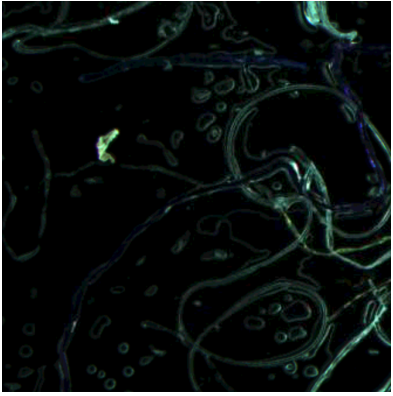

Fig 12: Complex fiber pattern

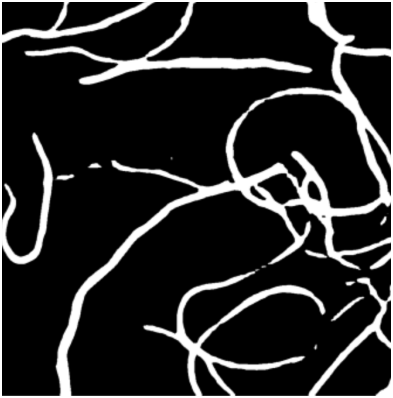

Fig. 13: Associated segmentation mask

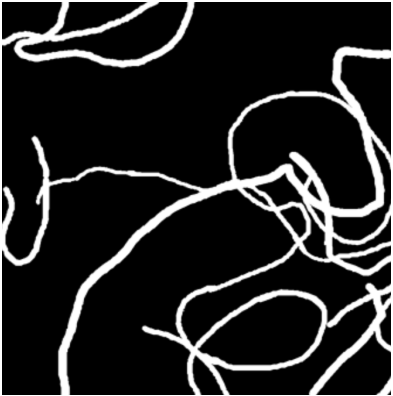

Fig. 14: Annotationimage

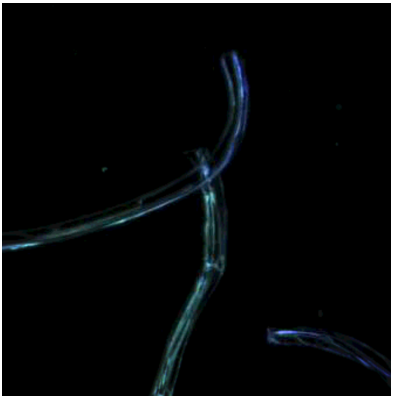

Fig. 15: Fiber pattern with completely different structure

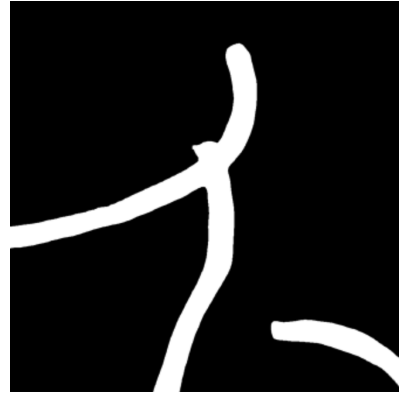

Fig. 16: Associated segmentation mask

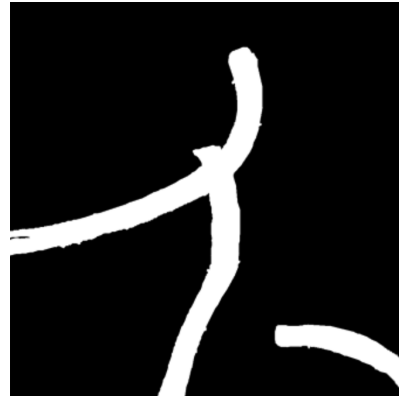

Abb. 17: Annotationimage

## V. FUSION OF MASK R-CNN INFERENCE WITH PSPNET

In order to combine the two model predictions, the following methods were used:

### A. Polygon Merging

By applying polygon merging, polygons are merged if they are on the same object. To detect this scenario, a calculation of the IoU (7) and a containment check is performed.

### B. Containment Check

If polygon A, which contains another polygon B in whole or in part, the IoU is small in this case, B is redundant and thus can be merged. For this scenario, we compute (17). If the containment exceeds the threshold, the polygon is merged.

$$C = \frac{TP}{(TP+FP)} \qquad (17)$$

So this procedure has now been used to connect the formed masks of Mask R-CNN and PSPNet.

The results are shown below (Fig. 17, Fig. 18). In order to ensure better comparability and clarity of the images, they have been designed as overlay images, i.e. the mask has been superimposed on the original image. Mask R-CNN shows the familiar image here (cf. Fig. 1), whereas a prediction connection yields significantly better results.

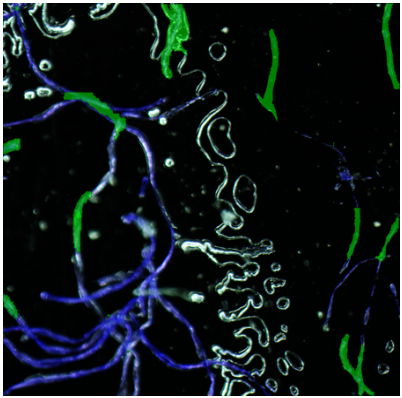

Fig. 18: Overlay Image with Mask R-CNN

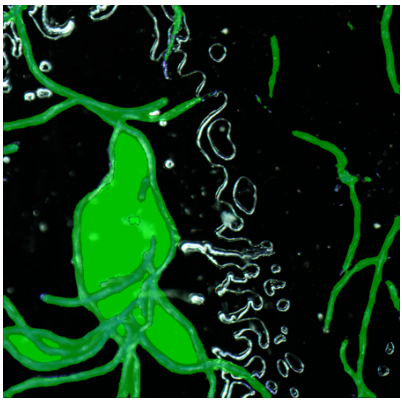

Fig. 19: Overlay image with Mask R-CNN and PSPNet, marked area is false positive

## VI. Conclusion

It could be shown that despite the challenging example, a strategy could be developed to solve the problem. Considering additionally that, for example, the pixel error in the ADE20K dataset by the expert annotator was 17.6% [18], a result above 80% meanIoU can be considered immensely good. In addition, we were able to combine the prediction masks of both models to obtain the optimum. Even though the segmentation was extremely challenging due to the great diversity within the class, we succeeded in achieving the best possible generalizability within PSPNet through a systematic approach to continuously improve the segmentation results. Our documented approach should help practitioners to improve their model quality for edge cases as well as encourage researchers to further develop Mask R-CNN and make it applicable for the segmentation of long, thin, diagonal objects.

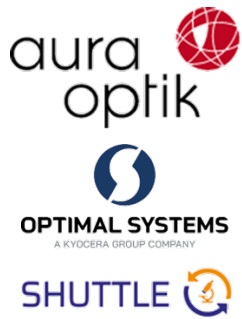

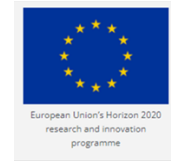

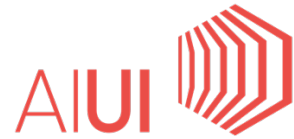

## VII. Links

*Informationen zum Projekt*:
https://www.shuttle-pcp.eu/

*Informationen zu den Partnern*:

Optimal Systems:
https://www.optimal-systems.de/

Aura Optik:
https://www.aura-optik.de/

AIUI:
https://www.ai-ui.ai/

## REFERENCES

[1] K. He, G. Gkioxari, P. Dollár, R. Girschick, "Mask R-Cnn", Facebook AI Researchgroup, 24.01.2018

[2] M. Elgendy, "Deep Learning for Vision Systems", Manning, 2020

[3] T. Akiba, S. Sano, T. Yanase, T. Ohta, M. Koyama, "Optuna: A next Generation Hyperparameter Optimization Framework", 25.07.2019

[4] S. Richter, "Statistisches und Maschinelles Lernen", Springer Spektrum, 2019

[5] C. Stein, C. Gausterer, "Gute und Schlechte DNA Spuren, ein Überblick zum aktuellen Stand der Molekularbiologie", SIAK Journal, 02.2017

[6] H. Zhao, J.Shi, X. Qi, X. Wang, J. Jia, „Pyramid Scene Parsing Network", 27.05.2017

[7] N. Srivastava, G. Hinton, A. Krizhevsky, I.Sutskever, R. Salakhutinov, "Dropout: A simple Way to prevent Neural Networks from Overfitting", Journal of Machine Learning Research, 06.2014.

[8] C. Shorten, T. M. Khoshgoftaar, „A Survery on Image Data Augmentation for Deep Learning", Journal of Big Data, 2019

[9] A. Gotmare, N.S. Keskar, C. Xiong, R.Socher, "A Closer Look at Deep Learning Heuristics: Learning Rate Restarts, Warmup and Destillation", 29.10.2018

[10] A. Senior, G. Heigold, M. A. Ranzato, K. Yang, "An Empirical Study of Learning Rates in Deep Neural Networks for Speech Recognition", IEEE International Conferences on Acoustics, Speech and Signale Processing, 05.2013

[11] R. Kleinberg, Y. Li, Y. Yuan, „An Alternative View: When does SGD Escape Local Minima?", 16.08.2018

[12] K. You, J. Wnag, M. Long, M. I. Jordan, "How does Learning Rate Decay Help Modern Neural Networks", 26.09.2019

[13] I. Goodfellow, Y. Bengio, A. Courville, "Deep Learning", MIT Press, 2016

[14] N.S. Keskar, D. Mudigere, J. Nocedal, M. Smelyanskiey, P.T.P Tang, „On Large Batch Training for Deep Learning: Generelization Gap and Sharp Minima".Conference Paper, ILCR 2017

[15] S. Jadon, "A Survey of Loss Functions for Semantic Segmentation", 03.09.2020

[16] H. Rezatofighi, N. Tsoi, J.Y. Gwak, A. Sadeghian, I. Reid, S. Savarese: "Generalized Intersection Over Union: A Metric and A Loss for Bounding Box Regression", 15.04.2019

[17] P. Ramachandran, B. Zoph, Q. V. Le: „Swish: A Self-Gated Activation Function", Google, 06.10.2017

[18] B. Zhou, H. Zhao, X. Puig, S. Fidler, A. Barriuso, A. Torralba, "Semantic understanding of scenes through the ADE20K dataset" 2016